\pdfoutput=1

\documentclass{article}
\usepackage{ijcai19}

\usepackage{times}
\usepackage{soul}
\usepackage{url}
\usepackage[utf8]{inputenc}
\usepackage[small]{caption}
\usepackage{graphicx}
\usepackage{amsmath}
\usepackage{booktabs}
\usepackage{algorithm}
\usepackage{algorithmic}
\usepackage{graphicx,times,amsmath} 
\usepackage{url}
\usepackage{amsfonts}
\usepackage{wasysym}
\usepackage{latexsym}
\usepackage{color}
\usepackage{multirow}
\usepackage{epstopdf}

\usepackage{array}
\usepackage{multirow}
\usepackage{verbatim}
\usepackage{amstext}
\usepackage{rotating}
\usepackage{amssymb}
\usepackage{subfigure}
\usepackage[svgnames]{xcolor}
\usepackage{float}
\usepackage{multirow}
\usepackage{colortbl}
\usepackage{caption}

\newcommand{\argmax}[1]{\underset{#1}{\operatorname{arg}\,\operatorname{max}}\;}

\usepackage{tikz}
\usetikzlibrary{positioning,fit,arrows.meta,backgrounds}
\usetikzlibrary{shapes,decorations,arrows}
\usetikzlibrary{calc}

\tikzset{
    module/.style={%
        draw, rounded corners,
        minimum width=#1,
        minimum height=7mm,
        font={\footnotesize}
        },
    module/.default=2cm,
   	beats/.style={thick,->,>=stealth',draw}
}

\title{Split Q Learning: Reinforcement Learning with Two-Stream Rewards}

\begin{document}


\author{
    Baihan Lin$^1$\and
    Djallel Bouneffouf$^2$\And 
    Guillermo Cecchi$^2$  
    \affiliations
    $^1$ Center for Theoretical Neuroscience, Columbia University, New York, NY 10027, USA \\
    $^2$ IBM Thomas J. Watson Research Center, Yorktown Heights, NY 10598, USA
    \emails
    Baihan.Lin@Columbia.edu, 
    \{dbouneffouf, gcecchi\}@us.ibm.com
}

\maketitle

\begin{abstract}

Drawing an inspiration from behavioral studies of human decision making, we propose here a general parametric framework for a reinforcement learning problem, which extends the standard Q-learning approach to incorporate a two-stream framework of reward processing with biases biologically associated with several neurological and psychiatric conditions, including Parkinson's and Alzheimer's diseases, attention-deficit/hyperactivity disorder (ADHD), addiction, and chronic pain. For AI community, the development of agents that react differently to different types of rewards can enable us to understand a wide spectrum of multi-agent interactions in complex real-world socioeconomic systems. Moreover, from the behavioral modeling perspective, our parametric framework can be viewed as a first step towards a unifying computational model capturing reward processing abnormalities across multiple mental conditions and user preferences in long-term recommendation systems.

\end{abstract}

\section{Introduction}

In order to better understand and model human decision-making behavior, scientists usually investigate reward processing mechanisms in healthy subjects \cite{perry2015reward}. However, neurodegenerative and psychiatric disorders, often associated with reward processing disruptions, can provide an additional resource for deeper understanding of human decision making mechanisms. Furthermore, from the perspective of evolutionary psychiatry, various mental disorders, including depression, anxiety, ADHD, addiction and even schizophrenia can be considered as ``extreme points'' in a continuous spectrum of behaviors and traits developed for various purposes during evolution, and  somewhat less extreme versions of those traits can be actually beneficial in specific environments (e.g., ADHD-like  fast-switching attention can be life-saving in certain environments, etc.). Thus, modeling decision-making biases and traits associated with various disorders may actually enrich the existing computational decision-making models, leading to potentially more flexible and better-performing algorithms.
 
Herein, we focus on reward-processing biases associated with several mental disorders, including Parkinson's and Alzheimer's disease, ADHD, addiction and chronic pain. Our questions are: is it possible to extend standard reinforcement learning algorithms to mimic human behavior in such disorders? Can such generalized approaches outperform standard reinforcement learning algorithms on specific tasks?

We show that both questions can be answered positively. We build upon the Q Learning, a state-of-art approach to RL problem, and extend it to a parametric version which allows to split the reward information into positive stream and negative stream with various reward-processing biases known to be associated with particular disorders. For example, it was shown that (unmedicated) patients with  Parkinson's disease appear to learn better from negative rather than from positive rewards \cite{frank2004carrot};  another example is addictive behaviors which may be associated with an inability to forget strong stimulus-response associations from the past, i.e. to properly discount past rewards \cite{redish2007reconciling}, and so on. \emph{More specifically, we propose a parametric model which introduces weights on incoming positive and negative rewards, and on reward histories, extending the standard parameter update rules in Q Learning; tuning the parameter settings allows us to better capture specific reward-processing biases}.

\section{Proposed Approach: Split Q Learning}
\label{subsec:SQL}

We propose Split Q Learning (SQL), outlined in Algorithm \ref{alg:SQL}, which updates the Q values using four weight parameters: $\phi_1 $ and $\phi_2$ are the weights of the previously accumulated positive and negative rewards, respectively, while $\phi_3$ and $\phi_4$ represent the weights on the positive and negative rewards. In our algorithm, we have two Q tables that we are using $Q^{+}$ and $Q^{-}$ which respectively record the positive and negative feedback.

\begin{algorithm}[H]
\small
 \caption{Split Q Learning (SQL)}
 \label{alg:SQL}
 \begin{algorithmic}[1]
 \STATE {\bfseries }\textbf{For} each episode $t$ \textbf{do}
 \STATE {\bfseries } \quad Initialize $s$
 \STATE {\bfseries } \quad \textbf{Repeat} 
 \STATE {\bfseries }  \quad \quad  $Q(s,a) := \phi_2 Q^{+}(s,a) + \phi_4 Q^{-}(s,a)$
 \STATE {\bfseries } \quad \quad action $i_t= \argmax{i} Q_i(t)$, observe $s'\in S$, $r \in R(s)$
 \STATE {\bfseries }  \quad \quad $Q^{+}(s,a):= \phi_1\hat{Q}^{+}(s,a)+\alpha_t(r^{+}+\gamma \hat{V}^{+}(s')-\hat{Q}^{+}(s,a))$
 \STATE {\bfseries } \quad \quad $Q^{-}(s,a):= \phi_3\hat{Q}^{-}(s,a)+\alpha_t(r^{-}+\gamma \hat{V}^{-}(s')-\hat{Q}^{-}(s,a))$
 \STATE {\bfseries } \quad \textbf{until} $s$ is terminal
 \end{algorithmic}
\end{algorithm}

\section{Reward Processing with Different Biases}

\begin{table}[tb]
\centering
\resizebox{0.8\columnwidth}{!}{
 \begin{tabular}{ l | c | c | c | c }
  & $\phi_1$   & $\phi_2$     & $\phi_3$      & $\phi_4$ \\ \hline
  AD (addiction)   & $1 \pm 0.1$   & $1 \pm 0.1$    & $0.5 \pm 0.1$   & $1 \pm 0.1$ \\
  ADHD  & $0.2\pm 0.1$  & $1 \pm 0.1$    & $0.2 \pm 0.1$   & $1 \pm 0.1$ \\
  AZ (Alzheimer's)   & $0.1 \pm 0.1$  & $1 \pm 0.1$    & $0.1 \pm 0.1$   & $1 \pm 0.1$ \\
  CP (chronic pain)   & $0.5 \pm 0.1$  & $0.5 \pm 0.1$    & $1 \pm 0.1$    & $1 \pm 0.1$ \\
  bvFTD   & $0.5 \pm 0.1$  & $100 \pm 10$    & $0.5 \pm 0.1$   & $1 \pm 0.1$ \\
  PD (Parkinson's)   & $0.5 \pm 0.1$  & $1 \pm 0.1$    & $0.5\pm 0.1$   & $100\pm 10$\\
 M (``moderate'')   & $0.5 \pm 0.1$  & $1 \pm 0.1$    & $0.5 \pm 0.1$   & $1 \pm 0.1$ \\
  standard SQL  & 1        & 1         & 1         & 1\\
 \end{tabular}
 }
 \caption{Algorithm Parameters}
\label{tab:parameter}
 \label{table:Synthetic}
 \end{table}
 
In this section we describe how specific constraints on the model parameters in the proposed algorithm can yield different reward processing biases, and introduce several instances of the SQL model, with parameter settings reflecting particular biases. The parameter settings are summarized in Table \ref{tab:parameter} as elaborated in \cite{bouneffouf2017bandit}.
Of course, the above models should be treated only as first approximations of the reward processing  biases in mental disorders, since the actual changes in reward processing are much more complicated, and the parameteric setting must be learned from actual patient data.
Herein, we first consider those models as specific variations of our general method, inspired by certain aspects of the corresponding diseases, and focus primarily on the computational aspects of our algorithm, demonstrating that the proposed parametric extension of Q Learning can learn better than the baseline Q Learning due to added flexibility. In the second step of this research, we utilize inverse reinforcement learning (IRL) \cite{abbeel2004apprenticeship} to learn the most likely reward function $\mathcal{R}_E$ of a human subject as an executing expert $E$ given collected behavioral trajectory consisting of a sequence of state-action pairs. Given a weight vector $\boldsymbol{w}$, one can compute the optimal policy $\pi_{\boldsymbol{w}}$ for the corresponding reward function $\widehat{\mathcal{R}}_{\boldsymbol{w}}$, and estimate its feature expectations $\hat{\mu}(\pi_{\boldsymbol{w}})$. IRL compares this $\hat{\mu}(\pi_{\boldsymbol{w}})$ with expert's feature expectations $\hat{\mu}_E$ to learn best fitting weight vectors $\boldsymbol{w}$. Instead of a single weight vector, the IRL algorithm learns a set of possible weight vectors, and they ask the agent designer to pick the most appropriate weight vector among these by inspecting their corresponding policies. In this way, we learn the parameters $\phi_1$, $\phi_2$, $\phi_3$, $\phi_4$ for the human subjects.

\section{Reward-Scaling in Reinforcement Learning}

To demonstrate the computational advantage of our proposed two-stream parametric extension of Q Learning can learn better than the baseline Q Learning, we tested our agents in nine computer games: Pacman, Catcher, FlappyBird, Pixelcopter, Pong, PuckWorld, Snake, WaterWorld, and Monster Kong. In each game, we tested in both stationary and non-stationary environments by rescaling the size and frequency of the reward signals in two streams. Preliminary results suggest that SQL outperform classical Q Learning in the long term in certain conditions (for example, positive-only and normal reward environments in Pacman). Our results also suggests that SQL behaves differently in the transition of reward environments. To understand this discrepancy, we further developed a variant of SQL which updates its four bias parameters adaptively with the Gaussian Process Upper Confidence Bound (GP-UCB) algorithm \cite{srinivas2009gaussian}. 
 
\section{Conclusion and Future Work}
\label{sec:Conclusion}

This research proposes a novel parametric family of algorithms for RL problem, extending the classical Q Learning to model a wide range of potential reward processing biases. Our approach draws an inspiration from extensive literature on decision-making behavior in neurological and psychiatric disorders stemming from disturbances of the reward processing system, and demonstrates high flexibility of our multi-parameter model which allows to tune the weights on incoming two-stream rewards and memories about the prior reward history. Our preliminary results support multiple prior observations about reward processing biases in a range of mental disorders, thus indicating the potential of the proposed model and its future extensions to capture reward-processing aspects across various neurological and psychiatric conditions. 

The contribution of this research is two-fold: from the AI perspective, we propose a more powerful and adaptive approach to RL, outperforming state-of-art QL; from the neuroscience perspective, this work is the first attempt at general, unifying model of reward processing and its disruptions across a wide population including both healthy subjects and those with mental disorders, which has a potential to become a useful computational tool for neuroscientists and psychiatrists studying such disorders. Among the directions for future work, we plan to investigate the optimal parameters in a series of computer games evaluated on different criteria, for example, longest survival time vs. highest final score. In addition, we also plan to explore the multi-agent interactions given different reward processing bias. These discoveries can help build more interpretable real-world RL systems. On the neuroscience side, the next steps would include further tuning and extending the proposed model to better capture observations in modern literature, as well as testing the model on both healthy subjects and patients with specific mental conditions. 

\small
\bibliography{main}

\bibliographystyle{named}

\end{document}